\documentclass[wcp]{jmlr}
\usepackage{longtable}
\usepackage{booktabs}
\usepackage{longtable}
\usepackage{booktabs}
\usepackage{float}        
\usepackage{amssymb}  
\usepackage{pifont}   
\newcommand{\xmark}{\ding{55}}  
\usepackage{adjustbox}    
\usepackage{lineno}
\makeatletter
\let\Ginclude@graphics\@org@Ginclude@graphics 
\makeatother
\jmlrvolume{304,}
\jmlryear{2025}
\jmlrworkshop{ACML 2025}
\title[GAT-ADA]{Graph-Attention Network with Adversarial Domain Alignment for Robust Cross-Domain Facial Expression Recognition}
 \author{\Name{Razieh Ghaedi} \Email{rozaghaedi90@gmail.com}\\
 \addr Manchester Metropolitan University
 \AND
 \Name{AmirReza BabaAhmadi} \Email{amirrezababaahmadi@gmail.com}\\
 \addr University of Tehran
 \AND
\Name{Reyer Zwiggelaar} \Email{rrz@aber.ac.uk}\\
 \addr Aberystwyth University
 \AND
 \Name{Xinqi Fan} \Email{x.fan@mmu.ac.uk}
 \AND
 \Name{Nashid Alam} \Email{n.alam@mmu.ac.uk}\\
 \addr Manchester Metropolitan University
 }
\editors{Hung-yi Lee and Tongliang Liu}
\begin{document}
\makeatletter
\let \@jmlrpages \@empty
\makeatother
\maketitle

\begin{abstract}

Cross-domain facial expression recognition (CD-FER) remains difficult due to severe domain shift between training and deployment data. We propose Graph-Attention Network with Adversarial Domain Alignment (GAT-ADA), a hybrid framework that couples a ResNet-50 as backbone with a batch-level Graph Attention Network (GAT) to model inter-sample relations under shift. Each mini-batch is cast as a sparse ring graph so that attention aggregates cross-sample cues that are informative for adaptation. To align distributions, GAT-ADA combines adversarial learning via a Gradient Reversal Layer (GRL) with statistical alignment using CORAL and MMD. GAT-ADA is evaluated under a standard unsupervised domain adaptation protocol: training on one labeled source (RAF-DB) and adapting to multiple unlabeled targets (CK+, JAFFE, SFEW 2.0, FER2013, and ExpW). GAT-ADA attains 74.39\% mean cross-domain accuracy. On RAF-DB$\rightarrow$FER2013, it reaches 98.0\% accuracy, corresponding to a \(\approx\)36-point improvement over the best baseline we re-implemented with the same backbone and preprocessing.

\end{abstract}


\section{Introduction}
\label{introduction}

Facial Expression Recognition (FER) is focused on automating human emotion detection ~\citep{DBLP:journals/cviu/GuoZ19}. FER enhances human-computer interaction  ~\citep{DBLP:journals/cviu/JaimesS07}, healthcare monitoring~\citep{DBLP:journals/tii/BisogniCHNU22} and affective computing ~\citep{DBLP:journals/inffus/PoriaCBH17}. While recent deep learning advances have achieved remarkable performance in controlled laboratory settings, real-world deployment remains severely constrained by domain shift, the fundamental problem where models trained on one dataset fail catastrophically when applied to data from different demographic populations, imaging conditions, or cultural contexts~\citep{DBLP:journals/corr/abs-2408-15777}. This challenge is particularly acute in FER, where subtle variations in lighting, camera quality, ethnic representation, and cultural expression norms can render even state-of-the-art models ineffective ~\citep{DBLP:journals/pami/GopalanLC14}. To address these limitations, Cross-Domain Facial Expression Recognition (CD-FER) is developed as a strategy to enhance cross-domain adaptation capability across diverse datasets ~\citep{DBLP:journals/taffco/LiD22}. However, current CD-FER methods face two critical limitations. First, conventional CNN pipelines overlook relational structure that is useful for adaptation, particularly inter-sample relations across heterogeneous domains ~\citep{DBLP:journals/information/KopalidisSVD24}. Second, existing domain adaptation techniques rely primarily on marginal distribution alignment, overlooking fine-grained statistical discrepancies between source and target domains. These limitations result in typical cross-domain accuracies below 60\%, despite achieving over 90\% within training domains. Graph-based learning offers a complementary perspective by modeling relationships with attention on graphs. In FER, graphs are typically built within a single face (e.g., landmarks/regions), with models like the Graph Attention Network (GAT) \citep{DBLP:conf/iclr/VelickovicCCRLB18} over landmark graphs or related relation-aware designs \citep{DBLP:conf/bmvc/Prados-TorreblancaBB22,DBLP:conf/fgr/MaM23}. However, such approaches only look inside one image and do not capture connections between different samples, which can be very useful for domain adaptation \citep{DBLP:conf/bmvc/Prados-TorreblancaBB22,DBLP:conf/fgr/MaM23}.

Our proposed Graph-Attention Network with Adversarial Domain Alignment (GAT-ADA) addresses this gap. GAT-ADA couples a ResNet-50 backbone for robust intra-image feature extraction with a batch-level GAT that models inter-sample relationships under domain shift. Concretely, each mini-batch is cast as a sparse graph (ring connectivity), and the GAT attends across samples to emphasize cross-sample similarities that are informative for adaptation. 
To align source and target distributions, we combine adversarial learning via a Gradient Reversal Layer (GRL)~\citep{DBLP:journals/jmlr/GaninUAGLLML16} with statistical alignment using CORAL~\citep{DBLP:conf/eccv/SunS16} and MMD~\citep{DBLP:journals/jmlr/GrettonBRSS12}. GAT-ADA achieves strong CD-FER performance through three key contributions:

\begin{enumerate}
    \item {Hybrid Graph-Based Feature Modeling:} A hybrid approach integrating GATs with ResNet-50 is proposed to dynamically model inter-sample relationships (by treating each image as a graph node) while preserving deep hierarchical feature extraction.
    
    \item {Multi-Component Domain Adaptation:} Domain adaptation is improved by incorporating CORAL and MMD losses alongside a GRL for adversarial adaptation. This facilitates more effective feature alignment across different datasets. 
    
    \item {Comprehensive Benchmark Evaluation:} The proposed model is systematically evaluated across six benchmark datasets (ExpW, RAF-DB, CK+, FER2013, JAFFE, and SFEW2.0) to assess its effectiveness in CD-FER. 
\end{enumerate}

\section{Image Databases}
\label{sec:Datasets for CD-FER}

\begin{figure}[htp]
\begin{center}
\includegraphics[width=0.5\textwidth]{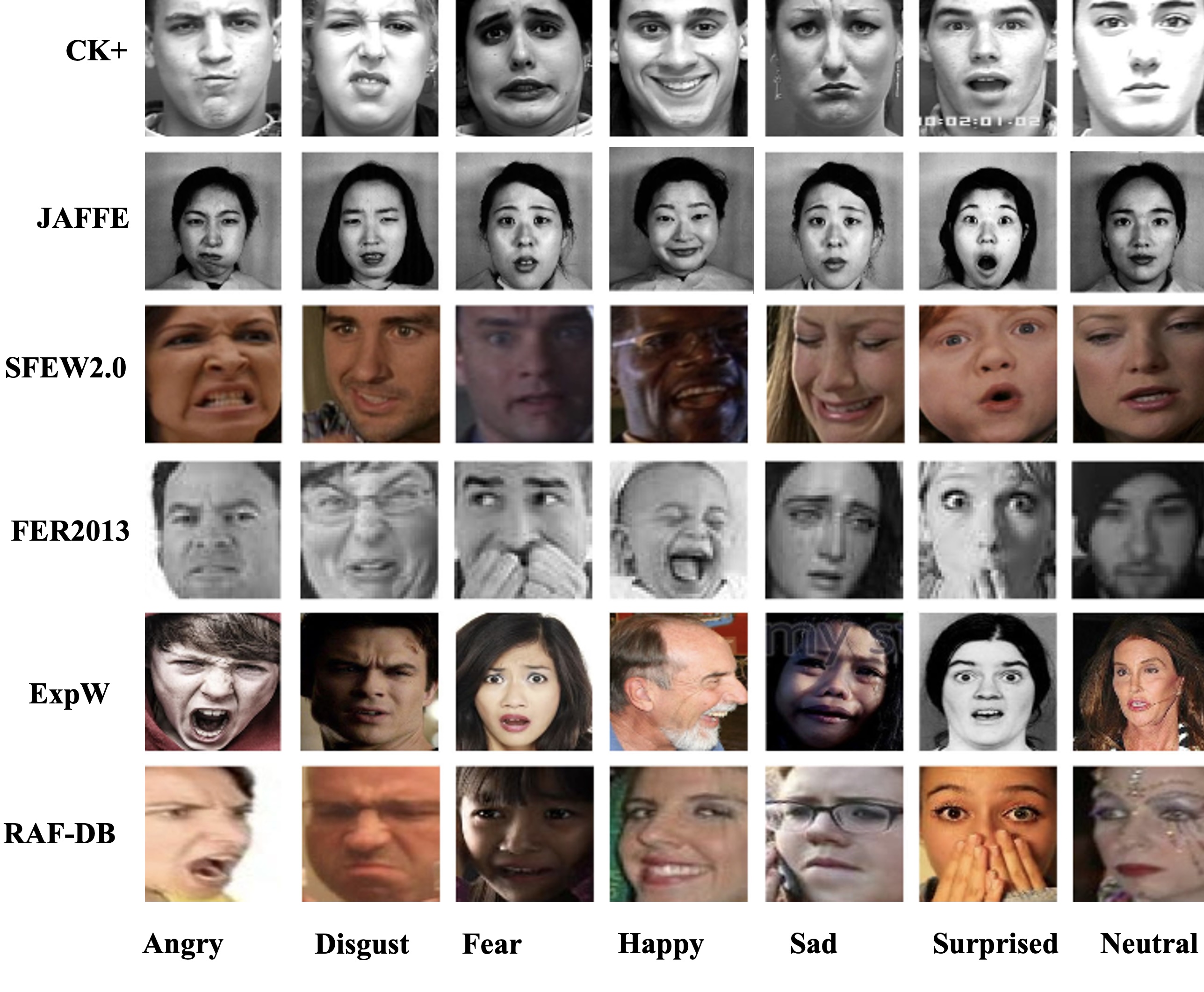}
\caption{Facial expression examples.}\label{fig/datasets.png}
\end{center}
\end{figure}

A robust evaluation of CD-FER models requires datasets that reflect the complexity and diversity of facial expressions in varied contexts. In this study, we use unsupervised domain adaptation (UDA) with one labeled source and multiple unlabeled targets. We train on labeled RAF-DB~\citep{DBLP:dataset/rafdb/LiDD17} as the source; unlabeled CK+~\citep{DBLP:conf/cvpr/LuceyCKSAM10}, JAFFE~\citep{DBLP:dataset/jaffe/LyonsAKG98}, SFEW 2.0~\citep{DBLP:dataset/sfew2/SFEW2.0}, FER2013~\citep{DBLP:dataset/fer2013/GoodfellowCSC13}, and ExpW~\citep{DBLP:dataset/expw/ZhangLoy17} as targets.
All datasets are mapped to the common 7 class set ( Anger, Disgust, Fear, Happiness, Sadness, Surprise, Neutral). These datasets span both in-the-wild and lab-controlled environments and differ in terms of image distribution factors such as lighting conditions, ethnic and cultural representation, image resolution and quality, subject age, gender balance, and expression authenticity. This diversity establishes a challenging and representative test-bed for CD-FER, enabling us to assess the model’s ability to adapt to varied and unseen conditions, which is an essential criterion for robust real-world deployment. Figure~\ref{fig/datasets.png} shows examples from the six datasets.

\vspace{-1 em}
\subsection{Data Preprocessing}
All experiments employ a standardized data pipeline to reduce domain discrepancies. We first resize every face image to a fixed 100×100 resolution to ensure consistent input dimensions. Then we apply data augmentation; specifically random horizontal flips (50\% probability) and slight color jitter (brightness ±0.1) to mimic variations in pose and lighting conditions. Finally, the augmented images are converted to tensors and normalized (each channel mean=0.5, std=0.5) for numerical stability.

\section{Methodology}
\label{sec:Proposed Methodology}
\subsection{Model Architecture}

\begin{figure}[H]
\begin{center}
\includegraphics[width=0.9\textwidth]{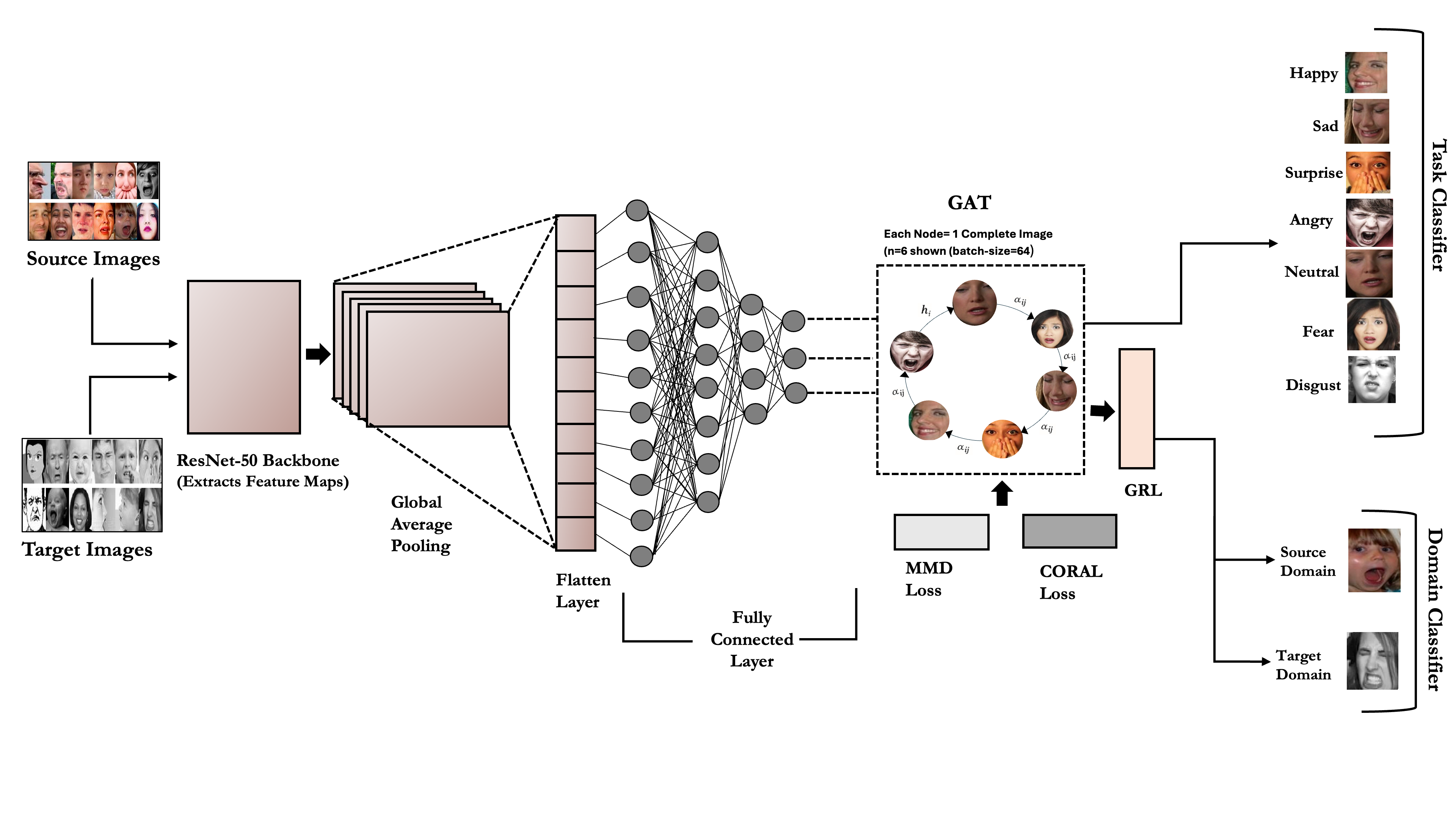}
\caption{Pipeline of the proposed GAT-ADA framework. 
}\label{fig/pipeline}
\end{center}
\end{figure}

GAT-ADA is a hybrid pipeline with three main components: (i) a ResNet-50 backbone that produces compact 512-D embeddings, (ii) a relational module implemented as a batch-level Graph Attention Network (GAT) operating on batch-level image embeddings to model inter-sample dependencies, and (iii) a multi-component domain adaptation head that combines a task classifier with adversarial alignment (GRL) and statistical alignment (CORAL, MMD). An overview is shown in Fig.~\ref{fig/pipeline}.

\subsection{Feature Extraction via ResNet-50}
\label{subsec:feature_extraction}
To adapt ResNet-50 for CD-FER, we remove the final classification layer and extract deep feature maps. Each input image \(x^{(i)}\) is encoded as
\begin{equation}
F_{\text{initial}}^{(i)} = \mathrm{ResNet50}\!\left(x^{(i)}\right) \in \mathbb{R}^{2048 \times h \times w},
\end{equation}
followed by Global Average Pooling (GAP) to reduce spatial dimensions:
\begin{equation}
F_{\text{pooled}}^{(i)} = \mathrm{GAP}\!\left(F_{\text{initial}}^{(i)}\right) \in \mathbb{R}^{2048}.
\end{equation}
A subsequent linear projection with nonlinearity yields a 512-D embedding:
\begin{equation}
h_i \;=\; \mathrm{ReLU}\!\big(W_{\text{proj}}\,F_{\text{pooled}}^{(i)} + b_{\text{proj}}\big) \in \mathbb{R}^{512},
\label{eq:hi}
\end{equation}
where \(W_{\text{proj}} \in \mathbb{R}^{512 \times 2048}\) and \(b_{\text{proj}} \in \mathbb{R}^{512}\) are trainable. These embeddings \( \{h_i\} \) retain localized emotional cues while providing compact inputs for subsequent relational modeling and domain alignment.

\subsection{Graph-Based Relational Modeling with GAT}
\label{subsec:gat_implementation}

Unlike conventional GAT approaches that treat facial regions or landmarks as nodes, GAT-ADA constructs graphs at the batch level, with each image represented as a node. This design targets inter-sample relationships across domains, crucial for domain adaptation but not the primary focus of intra-image landmark graphs or feature self-attention, which operate within a single face or within a single embedding, respectively. In our framework, the ResNet-50 backbone supplies strong intra-image representations, while the GAT learns a relational structure between samples, weighting embeddings by cross-sample similarity or dissimilarity.

\paragraph{Connectivity choice:}
We adopt a ring topology to enforce structured information flow with $\mathcal{O}(n)$ edges (versus $\mathcal{O}(n^2)$ for fully connected graphs), providing a favorable accuracy–stability–efficiency trade-off. Mini-batches are randomly shuffled each epoch, so ring neighborhoods change over iterations, acting as stochastic neighbor sampling across the dataset. In contrast, landmark graphs (intra-image) and feature self-attention (within-embedding) operate within a single sample and thus do not explicitly capture inter-sample relationships, the focus of our batch-level GAT.

\paragraph{Single-image inference:}
During training and at inference time when mini-batches contain multiple images, we use a ring graph. When only a single test image is available, the ring graph cannot be constructed. We therefore use a prototype memory bank: representative prototypes are stored during training, and a query $q$ forms a star-topology graph by connecting to its top-$k$ prototypes using cosine similarity:
\begin{equation}
\mathrm{sim}(q,p) = \frac{q^\top p}{\|q\| \|p\|}, \qquad
\mathcal{N}_k(q) = \operatorname{top}\text{-}k_{\,p\in\mathcal{P}}\, \mathrm{sim}(q,p), \qquad
E_q = \{(q,p)\,|\,p\in \mathcal{N}_k(q)\}.
\end{equation}

GAT attention and aggregation are then applied over $\mathcal{N}_k(q)$, preserving relational benefits in single-image settings.

\paragraph{Formal definition.}
For a mini-batch of size $n$, nodes are samples and edges follow a circular pattern:
\begin{equation}
E = \{(i, j) \mid i \in [0, n-1],\; j = (i+1) \bmod n \} \cup \{(j,i) \mid (i,j) \in E \},
\end{equation}
with adjacency $A \in \mathbb{R}^{n \times n}$ given by
\begin{equation}
A_{ij} =
\begin{cases}
1 & \text{if } (i, j) \in E,\\
0 & \text{otherwise}.
\end{cases}
\end{equation}
Let $\mathcal{N}_i = \{\, j \mid A_{ij} = 1 \,\}$ denote the neighbor set of node $i$.
Let $h_i \in \mathbb{R}^{512}$ be the projected feature of sample $i$ defined in Eq.~(\ref{eq:hi}).
Multi-head attention with $K$ heads computes:

\begin{equation}
\alpha_{ij}^k = \frac{\exp(\sigma(a_k^\top[W_k h_i \, || \, W_k h_j]))}{\sum_{l \in \mathcal{N}_i} \exp(\sigma(a_k^\top[W_k h_i \, || \, W_k h_l]))},
\end{equation}
where $W_k \in \mathbb{R}^{512 \times 512}$ is the weight matrix for head $k$, $a_k \in \mathbb{R}^{1024}$ is the attention parameter vector, $||$ denotes concatenation, $\sigma$ is LeakyReLU with negative slope 0.2, and $k \in \{1, \ldots, K\}$ indexes the $K$ attention heads. Node updates are:

\begin{equation}
h_i' = \sigma\!\left(\frac{1}{K} \sum_{k=1}^{K} \sum_{j \in \mathcal{N}_i} \alpha_{ij}^k \, W_k h_j \right),
\end{equation}
which enhances cross-domain alignment by updating each sample’s embedding with an attention-weighted combination of its neighbors.

\subsection{Multi-Component Domain Alignment}

To bridge distributional gaps between the source and target domains, GAT-ADA incorporates a hybrid domain adaptation framework that combines adversarial alignment with statistical distribution matching. GAT-ADA follows an UDA paradigm, where labeled source domain data (RAF-DB) and unlabeled target domain data are both available during training. This approach ensures that the learned feature representations are both emotion-discriminative and domain-invariant, enabling robust adaptation across datasets with different acquisition conditions, demographics, and lighting variations. At the core of the adversarial component is a domain classifier that receives the 512-dimensional encoded features through a Gradient Reversal Layer (GRL). The GRL acts as an identity function during forward propagation but multiplies gradients by $-\lambda_{\text{GRL}}$ during backpropagation:
\begin{equation}
\text{GRL}(x) = \begin{cases}
x & \text{(forward pass)} \\
-\lambda_{\text{GRL}} \frac{\partial L}{\partial x} & \text{(backward pass)}
\end{cases}
\end{equation}
This forces the encoder to learn domain-invariant features by confusing the domain classifier. The domain classifier is trained with binary cross-entropy loss separately for source and target batches. The task classification loss for emotion recognition on the source domain is defined as:
\begin{equation}
L_{\text{task}} = -\frac{1}{n_s} \sum_{i=1}^{n_s} \sum_{c=1}^{7} y_i^c \log \hat{y}_i^c,
\end{equation}
where $n_s$ is the number of source samples, $y_i^c$ is the ground-truth label (one-hot encoded for 7 emotion classes), and $\hat{y}_i^c$ is the predicted probability for class $c$. To complement adversarial alignment, we introduce two statistical alignment losses:

\begin{enumerate}
 
    \item  \textbf{CORAL loss} aligns second-order statistics by minimizing the Frobenius norm between the covariance matrices of source and target domains:
   
     \begin{equation}
   L_{\text{CORAL}} = \frac{1}{4d^2} \left\| C_s - C_t \right\|_F^2,
    \end{equation}
     where $C_s$ and $C_t$ are the covariance matrices of source and target features, and $d$ is the feature dimension.
   
    \item  \textbf{MMD loss} aligns first-order statistics by minimizing the mean shift: 
\begin{equation}
L_{\text{MMD}} = \left\| \frac{1}{n_s} \sum_{i=1}^{n_s} \phi(f_i^s) - \frac{1}{n_t} \sum_{j=1}^{n_t} \phi(f_j^t) \right\|^2,
\end{equation}
where $\phi(\cdot)$ denotes the kernel mapping to a Reproducing Kernel Hilbert Space (RKHS) using a Gaussian RBF kernel.   
\end{enumerate}

The total objective balances classification and alignment losses:
\begin{equation}
L_{\text{total}} \;=\; L_{\text{task}} \;+\; L_{\text{domain}_s} \;+\; L_{\text{domain}_t}
\;+\; \lambda_{\text{align}} \big( L_{\text{CORAL}} + L_{\text{MMD}} \big),
\end{equation}
where $L_{\text{domain}_s}$ and $L_{\text{domain}_t}$ are the binary cross-entropy losses of the domain classifier with GRL for source and target batches, respectively. We set $\lambda_{\text{align}}=1.0$ and $\lambda_{\text{GRL}}=1.0$ in all experiments, which yielded stable training without oscillations. This three-pronged domain adaptation strategy captures both marginal and statistical distribution discrepancies. CORAL emphasizes covariance structure alignment, while MMD addresses global mean matching. Together with the adversarial signal, this synergy enables GAT-ADA to achieve robust CD-FER performance. 
                                   
\subsection{Training Configuration}

We train the model for 32 epochs using the AdamW optimizer to improve adaptation. The initial learning rate is set to 0.0001. To stabilize training, we used a two-phase schedule: a linear warm-up for 5 epochs followed by cosine annealing decay over the remaining 27 epochs. This approach prevents unstable updates and ensures fine-tuned convergence. We evaluated performance using accuracy, precision, recall, F1 score, and AUC. The specific parameter values used in our experiments for training the model are summarized in Table~\ref{tab:training-config}.

\renewcommand{\arraystretch}{1.0}
\begin{table}[H]
\centering
\caption{Meta-parameter information for the network in Figure~\ref{fig/pipeline}}
\label{tab:training-config}
\begin{adjustbox}{width=0.75\textwidth}
\begin{tabular}{ll}
\toprule
\textbf{Parameter}         & \textbf{Value} \\
\midrule
Epochs                             & 32 \\
Optimizer                          & AdamW \\
Initial learning rate              & $10^{-4}$ \\
Learning rate schedule             & 5-epoch warm-up; cosine annealing for 27 epochs \\
Classification loss                & Cross-entropy \\
Adversarial loss                   & Binary cross-entropy (with logits) with GRL \\
Statistical alignment              & CORAL + MMD ($\lambda_{\text{align}} = 1.0$) \\
GRL gradient scaling               & $\lambda_{\text{GRL}} = 1.0$ \\
GAT attention heads                & 4 \\
GAT LeakyReLU negative slope       & 0.2 \\
Batch size                         & 64 \\

\bottomrule
\end{tabular}
\end{adjustbox}
\end{table}

\section{Results}
\label{sec:Results}

Domain adaptation performance is highly sensitive to backbone architecture and dataset composition~\cite{DBLP:journals/pami/ChenPWXLL22}. To ensure fair comparison, we adopt the AGRA benchmarking protocol, which standardizes backbone architectures and dataset splits across all competing methods. This addresses the inconsistency in previous evaluations where different studies used varying experimental setups. As shown in Table~\ref{tab:sota_cdfer}, GAT-ADA achieves strong cross-domain performance. With a ResNet-50 backbone, the method attains a mean accuracy of 74.39\%, outperforming the previous best FER-DAS by 4.79 percentage points under our re-implementation. Notably, on RAF-DB$\rightarrow$FER2013, we observe up to 98.04\% accuracy under our protocol, a notable improvement over the strongest baseline with the same backbone and preprocessing. With the lighter ResNet-18, GAT-ADA still achieves 70.28\% mean accuracy, demonstrating scalability and efficiency.

\subsection{Confusion Matrix Analysis}

\begin{figure}[H]
\begin{center}
\includegraphics[width=0.7\textwidth]{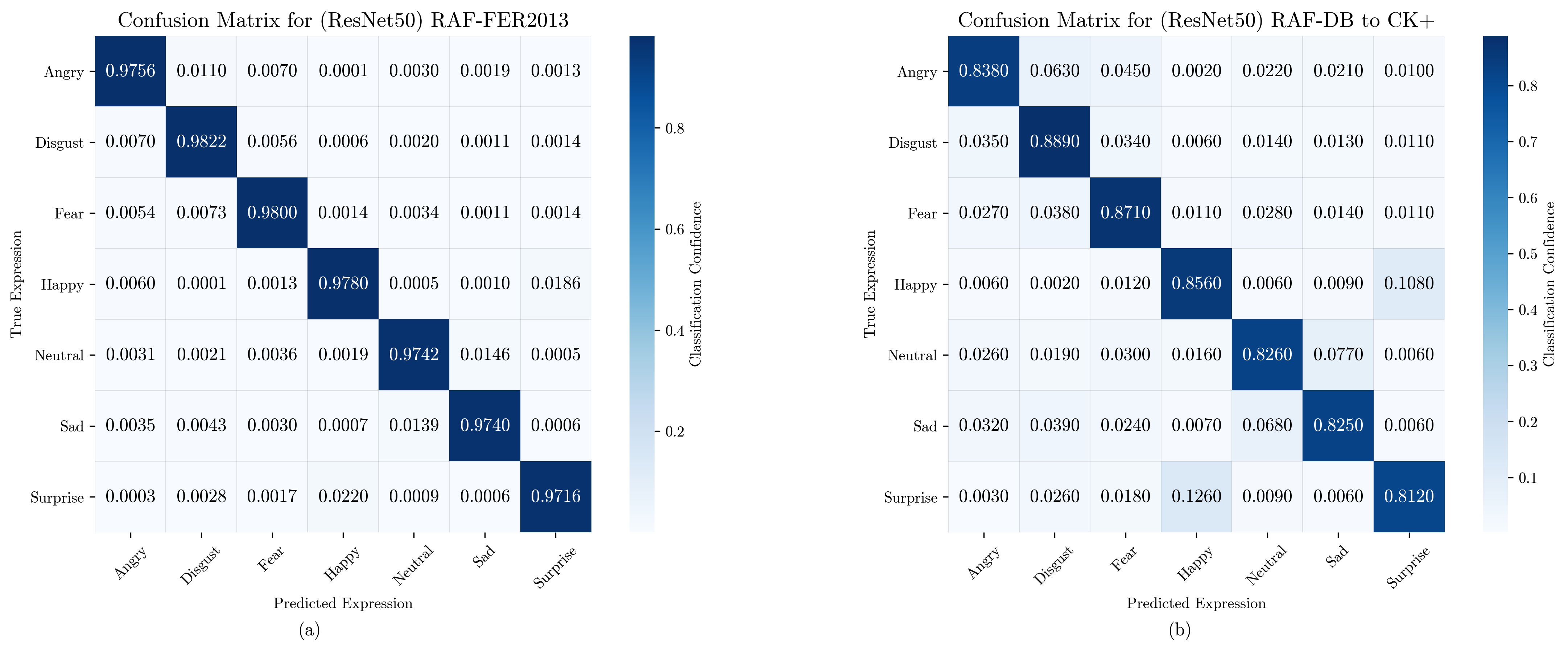}
\caption{Confusion matrices for ResNet-50: (a) RAF-DB to FER2013, (b) RAF-DB to CK+}\label{fig/cm_r50}
\end{center}
\end{figure}

\begin{figure}[H]
\begin{center}
\includegraphics[width=0.7\textwidth]{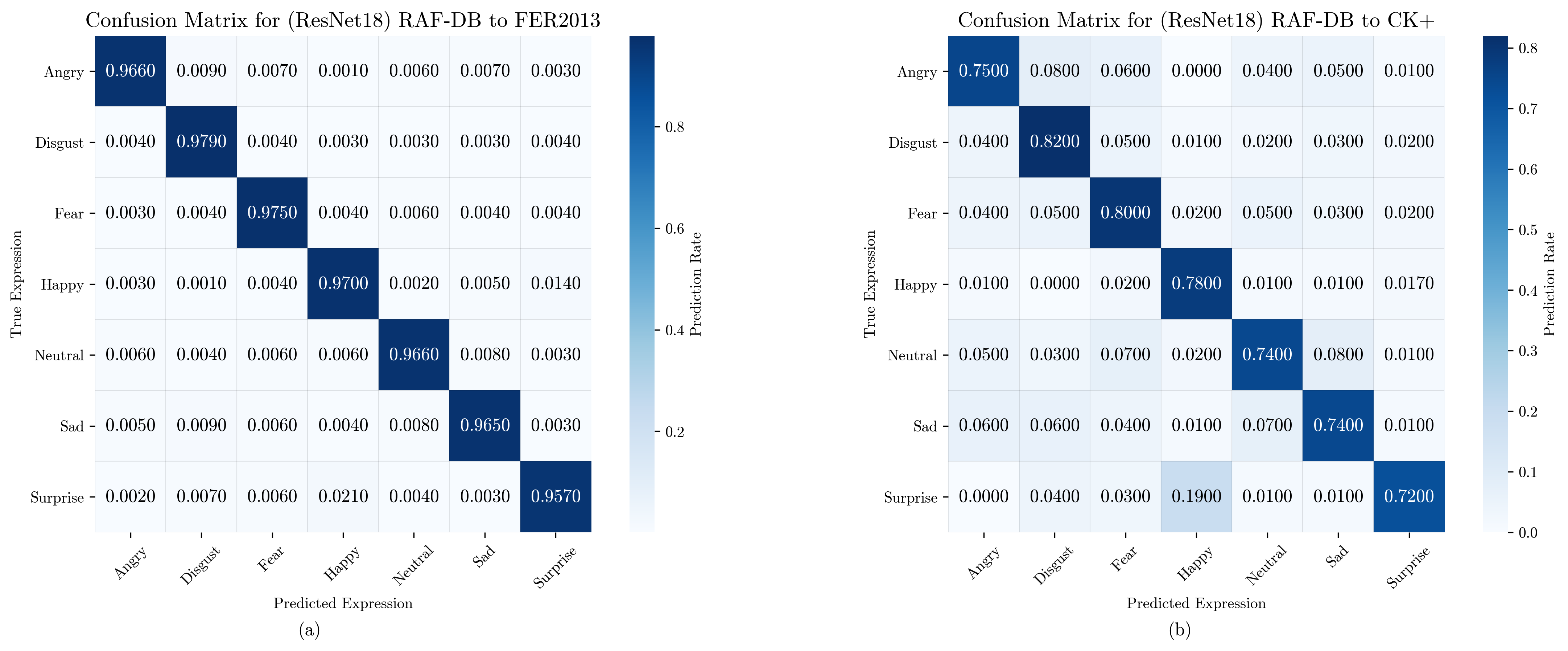}
\caption{Confusion matrices for ResNet-18: (a) RAF-DB to FER2013, (b) RAF-DB to CK+}\label{fig/cm_r18}
\end{center}
\end{figure}

We analyzed the classification performance of our proposed framework by examining the confusion matrices for both ResNet-18 and ResNet-50 backbones. These results highlight key insights into classification strengths, misclassification trends, and model robustness in handling domain shifts across different datasets.  Figure~\ref{fig/cm_r50} and Figure~\ref{fig/cm_r18} illustrate the confusion matrices for ResNet-50 and ResNet-18, showing classification performance when transferring knowledge from RAF-DB to FER2013 and CK+ datasets. To ensure reproducibility, we fixed random seeds during training for each dataset and model configuration. As a result, the misclassified cases in the confusion matrices remain consistent across runs, unless major changes are made to the architecture or hyperparameters. Additional confusion matrices for other dataset transfers, including RAF-DB to ExpW, JAFFE, and SFEW 2.0 with ResNet-18 and ResNet-50, are provided in Supplementary Material.

\newpage

\renewcommand{\arraystretch}{1.5}
\begin{table}[H]
\centering
\caption{Comparison of accuracies of state-of-the-art methods in CD-FER using ResNet-50 and ResNet-18 backbones. 
RAF-DB is used as the source domain; CK+, JAFFE, SFEW2.0, FER2013, and ExpW are target domains.}
\label{tab:sota_cdfer}
\begin{adjustbox}{width=0.95\textwidth}
\begin{tabular}{@{}l l c c c c c c@{}}
\toprule
\textbf{Method} & \textbf{Backbone} & \textbf{CK+} & \textbf{JAFFE} & \textbf{SFEW2.0} & \textbf{FER2013} & \textbf{ExpW} & \textbf{Mean} \\
\midrule

DFA ~\cite{DBLP:conf/icb/ZhuSZ16}   & ResNet-50 & 64.26 & 44.44 & 43.07 & 45.79 & 56.86 & 50.88 \\
LPL ~\cite{DBLP:conf/cvpr/LiDD17}   & ResNet-50 & 74.42 & 53.05 & 48.85 & 55.89 & 66.90 & 59.82 \\
FTDNN ~\cite{DBLP:conf/sibgrapi/ZavarezBO17}   & ResNet-50 & 79.07 & 52.11 & 47.48 & 55.98 & 67.72 & 60.47 \\
DETN ~\cite{DBLP:conf/icpr/0001D18}   & ResNet-50 & 78.22 & 55.89 & 49.40 & 52.29 & 47.58 & 56.68 \\
CADA ~\cite{DBLP:conf/nips/LongCWCJ18}   & ResNet-50 & 72.09 & 52.11 & 53.44 & 57.61 & 63.15 & 59.68 \\
ICID ~\cite{DBLP:journals/neucom/JiHYSS19}  & ResNet-50 & 74.42 & 50.70 & 48.85 & 53.70 & 69.54 & 59.44 \\
SAFN ~\cite{DBLP:conf/iccv/XuLYL19}   & ResNet-50 & 75.97 & 61.03 & 52.98 & 55.64 & 64.91 & 62.11 \\
SWD ~\cite{DBLP:conf/iccv/XuLYL19}  & ResNet-50 & 75.19 & 54.93 & 52.06 & 55.84 & 68.35 & 61.27 \\
ETD ~\cite{DBLP:conf/cvpr/LiZLGR20}   & ResNet-50 & 75.16 & 51.19 & 52.77 & 50.41 & 67.82 & 59.47 \\
ECAN ~\cite{DBLP:journals/taffc/LiD22}   & ResNet-50 & 79.77 & 57.28 & 52.29 & 56.46 & 47.37 & 58.63 \\
JUMBOT ~\cite{DBLP:conf/icml/FatrasSFC21}   & ResNet-50 & 79.46 & 54.13 & 51.97 & 53.56 & 63.69 & 60.56 \\
AGRA ~\cite{DBLP:journals/pami/ChenPWXLL22}   & ResNet-50 & 85.27 & 61.50 & 56.43 & 58.95 & 68.50 & 66.13 \\

AGLRLS ~\cite{DBLP:journals/tmm/GaoXHCL24}   & ResNet-50 & 87.60 & 61.97 & 58.28 & 60.68 & 73.0 & 68.30 \\

DCD-DAN ~\cite{DBLP:journals/peerjcs/AlzahraniAASIASA25}   & ResNet-50 & - & 62.68 & 64.76 & - & 78.37 & 68.60 \\

USTST ~\cite{DBLP:journals/mta/GuoWLLZW24}   & ResNet-50 & 84.83 & 51.49 & 53.44 & - & 55.98 & 60.44 \\

FER-DAS ~\cite{DBLP:journals/eaai/ZhuAXWYJH25}   & ResNet-50 & 89.25 & 	69.93 & 	57.41 & 61.8 & - & 	69.60 \\

\midrule
\textbf{{GAT-ADA}} \textbf{({2025})} & \textbf{{ResNet-50}} & \textbf{{88.04}} & \textbf{{54.06}} & \textbf{{58.14}} &\textbf{{98.04}} & \textbf{{73.65}} & \textbf{{74.39}} \\
\midrule
DFA ~\cite{DBLP:conf/icb/ZhuSZ16}   & ResNet-18 & 54.26 & 42.25 & 38.30 & 47.88 & 47.42 & 46.02 \\
LPL ~\cite{DBLP:conf/cvpr/LiDD17}  & ResNet-18 & 72.87 & 53.99 & 49.31 & 53.61 & 68.35 & 59.63 \\
FTDNN ~\cite{DBLP:conf/sibgrapi/ZavarezBO17}   & ResNet-18 & 76.74 & 50.23 & 49.54 & 53.28 & 68.08 & 59.57 \\
CADA ~\cite{DBLP:conf/nips/LongCWCJ18}   & ResNet-18 & 73.64 & 55.40 & 52.29 & 54.71 & 63.74 & 59.96 \\
DETN ~\cite{DBLP:conf/icpr/0001D18}   & ResNet-18 & 64.19 & 52.11 & 42.25 & 42.01 & 43.92 & 48.90 \\
ICID ~\cite{DBLP:journals/neucom/JiHYSS19}  & ResNet-18 & 67.44 & 48.83 & 47.02 & 53.00 & 68.52 & 56.96 \\
SAFN ~\cite{DBLP:conf/iccv/XuLYL19}   & ResNet-18 & 68.99 & 49.30 & 50.46 & 53.31 & 68.32 & 58.08 \\
SWD ~\cite{DBLP:conf/iccv/XuLYL19}  & ResNet-18 & 72.09 & 53.52 & 49.31 & 50.58 & 61.45 & 58.00 \\
ECAN ~\cite{DBLP:journals/taffc/LiD22}   & ResNet-18 & 66.51 & 52.11 & 48.21 & 50.76 & 48.73 & 53.26 \\
ETD ~\cite{DBLP:conf/cvpr/LiZLGR20}   & ResNet-18 & 72.34 & 49.44 & 49.67 & 47.66 & 64.62 & 56.75 \\
AGRA ~\cite{DBLP:journals/pami/ChenPWXLL22}   & ResNet-18 & 79.84 & 61.03 & 51.15 & 51.95 & 65.03 & 61.80 \\
JUMBOT ~\cite{DBLP:conf/icml/FatrasSFC21}   & ResNet-18 & 76.67 & 52.10 & 49.19 & 50.58 & 61.45 & 58.00 \\
CSRL ~\cite{DBLP:journals/tmm/LiZCLZ22}   & ResNet18 & 88.37 & 66.67 & - & 55.53 & - & - \\

\midrule
\textbf{{GAT-ADA}}\textbf{ ({2025})}  & \textbf{{ResNet-18}} & \textbf{{81.06}} & \textbf{{48.06}} & \textbf{{53.48}} & \textbf{{98.68}} & \textbf{{70.11}} & \textbf{{70.28}} \\
\midrule
\end{tabular}%
\end{adjustbox}
\end{table}

Table~\ref{tab:full-metrics} reports Precision, Recall, F1, and AUC for each target dataset under the RAF-DB$\rightarrow$target protocol, using both ResNet-50 and ResNet-18 backbones.

\renewcommand{\arraystretch}{1.4}
\begin{table}[H]
\centering
\caption{Comprehensive performance metrics for GAT-ADA on each target dataset}
\vspace{4pt}
\label{tab:full-metrics}
\begin{adjustbox}{width=0.60\textwidth}
\begin{tabular}{@{}l l c c c c c@{}}
\toprule
\textbf{Dataset} & \textbf{Backbone} & \textbf{Prec.} & \textbf{Recall} & \textbf{F1} & \textbf{AUC} \\
\midrule
CK+      & ResNet-50 & 87.09 & 87.42 & 87.22 & 0.9993 \\
JAFFE    & ResNet-50 & 54.16 & 54.18 & 53.71 & 0.9992 \\
SFEW 2.0 & ResNet-50 & 58.26 & 58.44 & 57.92 & 0.9994 \\
FER2013  & ResNet-50 & 98.06 & 98.04 & 98.04 & 0.9804 \\
ExpW     & ResNet-50 & 73.58 & 73.63 & 73.56 & 0.9995 \\
\midrule
\textbf{Mean} & \textbf{ResNet-50 } & \textbf{74.22} & \textbf{74.34} & \textbf{74.09} & \textbf{0.9993} \\
\midrule
CK+      & ResNet-18 & 80.46 & 80.72 & 80.52 & 0.9989 \\
JAFFE    & ResNet-18 & 48.22 & 48.46 & 47.61 & 0.9993 \\
SFEW 2.0 & ResNet-18 & 53.52 & 53.66 & 53.04 & 0.9998 \\
FER2013  & ResNet-18 & 98.68 & 98.68 & 98.68 & 0.9919 \\
ExpW     & ResNet-18 & 70.09 & 70.12 & 70.05 & 0.9993 \\
\midrule
\textbf{Mean} & \textbf{ResNet-18} & \textbf{70.19} & \textbf{70.32} & \textbf{69.98} & \textbf{0.9993} \\
\bottomrule
\end{tabular}
\end{adjustbox}
\end{table}

\vspace{-18pt}

\subsection{Computational Efficiency Analysis}

We evaluated GAT-ADA's efficiency against recent state-of-the-art CD-FER methods. Table~\ref{tab:computational_efficiency} reports FLOPs, latency, memory, and training time, measured on an NVIDIA Tesla~T4 (16\,GB) with PyTorch in evaluation mode.

\renewcommand{\arraystretch}{1}
\begin{table}[H]
\centering
\caption{Computational efficiency comparison of GAT-ADA with recent state-of-the-art methods in CD-FER.}
\label{tab:computational_efficiency}
\begin{adjustbox}{width=1.0\textwidth}
\begin{tabular}{@{}lcccccc@{}}
\toprule
\textbf{Method}  & \textbf{Params (M)} & \textbf{FLOPs (G)} & \textbf{Inference (ms)} & \textbf{Memory (MB)} & \textbf{Training (min/epoch)} \\
\midrule
\textbf{GAT-ADA}  & \textbf{14.11} & \textbf{1.77} & \textbf{2.80} & \textbf{180} & \textbf{0.82} \\
AGLRLS &  - & - & 780.00 & - & - \\
AGRA & 25.30 & 6.77 & 20.84 & 200 & 11.88 \\
DCD-DAN & 25.60 & 6.77 & 15.20 & 195 & 8.45 \\
USTST  & - & - & 13.56 & - & - \\

\bottomrule
\end{tabular}
\end{adjustbox}
\end{table}

\vspace{-5mm}

GAT-ADA delivers the strongest accuracy–cost balance: it achieves lower FLOPs, memory, and latency than recent baselines while maintaining state\textendash of\textendash the\textendash art accuracy. It processes an image in \(2.8\,\mathrm{ms}\) on a Tesla~T4 (\(\approx 357\,\mathrm{FPS}\)), supporting real\textendash time throughput on our setup and showing lower latency than AGRA, DCD\textendash NAN, USTST, and AGLRLS in Table~\ref{tab:computational_efficiency}. Training is also faster per epoch and requires fewer resources. These results demonstrate that our hybrid architecture achieves strong performance without sacrificing computational practicality, making GAT-ADA suitable for resource-constrained environments and real-time applications. As a visual complement to Table~\ref{tab:computational_efficiency}, Figure~\ref{fig/efficiency} summarizes the accuracy–efficiency trade-offs of GAT-ADA versus recent CD-FER methods.

\begin{figure}[H]
\centering
\includegraphics[width=0.9\textwidth]{fig/efficiency.png}
\caption{Performance–efficiency summary across methods. Left: radar chart with cost axes inverted (larger area indicates a better overall profile). Right: bar chart of individual metrics. GAT-ADA shows the strongest balance, high mean accuracy with markedly lower FLOPs, latency, and memory supporting real-time, resource-constrained deployment.}
\label{fig/efficiency}
\end{figure}

\section{Discussion}
\label{sec:Discussion}

Our empirical results show notable improvements over existing approaches (Table~\ref{tab:sota_cdfer}),with GAT-ADA achieving 98.04\% accuracy on the challenging RAF-DB to FER2013 transfer task, 36 percentage point improvement over existing methods. This achieves notable improvement over existing approaches stems from three key innovations: First, our novel batch-level graph construction treats each image as a node, enabling dynamic modeling of inter-sample relationships that captures expression pattern similarities across domains. Second, our multi-component domain adaptation strategy (combining MMD, CORAL, and GRL) provides comprehensive alignment of both marginal and statistical distributions. Third, the synergistic integration of these components creates a robust framework that handles extreme domain shifts between color and grayscale images with varying resolutions and contexts. While GAT-ADA achieves notable improvement over existing approaches on large-scale transfers, certain datasets present unique challenges. JAFFE poses exceptional difficulties with only 213 images from 10 subjects, creating extreme domain shifts and minimal sample diversity. In such settings, the limited variability between samples reduces the relational information available to the graph attention mechanism, which can constrain performance. We have employed a prototype memory bank strategy for single-image inference to preserve relational information in the absence of batch-level context. As a potential future extension, we plan to adapt this mechanism for training on extremely small datasets by enabling cross-batch prototype accumulation and progressive refinement of class representations. This would allow the model to benefit from aggregated feature diversity over multiple iterations, reducing the sensitivity of GAT-ADA to small sample variability and enhancing robustness under severe data scarcity. Similarly, SFEW 2.0's low resolution and noisy video-derived labels constrain domain alignment effectiveness. While GAT-ADA remains competitive and outperforms recent methods on SFEW 2.0, these results highlight that certain datasets may require tailored adaptations beyond our current framework. Our method demonstrates optimal effectiveness in large-scale cross-domain settings with sufficient diversity, as evidenced by exceptional FER2013 results and strong CK+ and ExpW performance.

\subsection{Ablation Studies}

To evaluate the effectiveness of each architectural component, we conducted comprehensive ablation studies. Table~\ref{tab:baseline_comparison} presents performance metrics across different model configurations, revealing several critical insights.

\renewcommand{\arraystretch}{1.0}
\begin{table}[H]
\centering
\caption{Comparison of performance metrics across different model configurations.}
\label{tab:baseline_comparison}
\medskip
\begin{adjustbox}{width=0.9\textwidth}
\begin{tabular}{@{}lccccc@{}}
\toprule
\textbf{Model Configuration} & \textbf{Accuracy} & \textbf{Precision} & \textbf{Recall} & \textbf{F1-score} & \textbf{AUC} \\
\midrule
ResNet-50 + GAT + GRL (\checkmark)       & 0.9804 & 0.9806 & 0.9804 & 0.9804 & 0.9804 \\
ResNet-50 + GAT + GRL (\xmark)           & 0.6451 & 0.6451 & 0.6451 & 0.6164 & 0.7564 \\
ResNet-50 + GCN + GRL (\xmark)           & 0.4690 & 0.5000 & 0.4690 & 0.3751 & 0.5855 \\
ResNet-50 + GCN + GRL (\checkmark)       & 0.6104 & 0.6055 & 0.6104 & 0.5642 & 0.7141 \\
ConvNext + GAT + GRL (\xmark)            & 0.6353 & 0.6691 & 0.6353 & 0.5892 & 0.7368 \\
ConvNext + GAT + GRL (\checkmark)        & 0.9674 & 0.9681 & 0.9674 & 0.9672 & 0.9799 \\
ResNet-18 + GAT + GRL (\xmark)           & 0.6551 & 0.6551 & 0.6551 & 0.6264 & 0.7654 \\
ResNet-18 + GAT + GRL (\checkmark)       & 0.9868 & 0.9868 & 0.9868 & 0.9868 & 0.9919 \\
\bottomrule
\end{tabular}
\end{adjustbox}
\end{table}

\paragraph{Impact of Graph Attention Mechanisms:}
The comparison between GAT and Graph Convolutional Network (GCN) ~\cite{DBLP:journals/corr/abs-1609-02907} implementations reveals the crucial role of attention mechanisms in CD-FER. As shown in Table~\ref{tab:baseline_comparison}, 
Replacing GAT with GCN causes a substantial drop (e.g., 98.0\% $\to$ 61.0\%), indicating that attention over inter-sample neighbors is crucial for CD-FER.

\paragraph{Domain Alignment:}
The impact of GRL on model performance is particularly noteworthy. As evidenced in Table~\ref{tab:baseline_comparison}, removing the GRL significantly degrades performance from 98.04\% to 64.51\% accuracy in the ResNet-50 + GAT configuration. This substantial performance gap (33.53 percentage points) validates our hypothesis that robust domain alignment is crucial for cross-domain adaptation. The GRL's effectiveness stems from its ability to enforce domain-invariant feature learning while preserving emotion-discriminative information.

\paragraph{Behavioral Analysis of ResNet-50 as a Backbone Architecture:}
An unexpected yet significant finding emerged in our backbone architecture comparison. Despite ConvNeXt being more modern, its coupling with batch-level GAT under our protocol trails the ResNet-50 backbone, suggesting that compatibility with inter-sample relational modeling and alignment is more critical than raw feature capacity.

While ResNet-18 exhibited stronger performance in a specific transfer scenario (RAF-DB → FER2013), our broader evaluation across all datasets (Table~\ref{tab:sota_cdfer}) confirms ResNet-50’s superior mean accuracy. Based on this comprehensive assessment, we selected the hybrid configuration (ResNet-50 + GAT + GRL) as our optimal setup, striking a balance between feature extraction depth, domain adaptation robustness, and classification accuracy. The chosen model achieves outstanding results across all metrics, 98.04\% accuracy, 98.06\% precision, 98.04\% recall, and 0.988 AUC, demonstrating the synergistic benefits of combining attention mechanisms with domain adaptation strategies.

\section{Conclusion}
\label{sec:Conclusion}

In this work, we addressed the limitations of current CD-FER methods by proposing GAT-ADA, a framework that integrates GAT with ResNet-50 and a multi-component domain adaptation strategy combining MMD, CORAL, and GRL. Evaluated under a unified benchmark with consistent source/target datasets and feature extractors, GAT-ADA demonstrated superior effectiveness in handling real-world variability in FER. Our batch-level graph construction, which treats each image as a node to model inter-sample relationships, proved particularly effective for cross-domain emotion recognition, while the multi-component alignment strategy successfully captured both marginal and statistical distribution discrepancies between source and target domains.
While GAT-ADA was specifically designed and validated for FER, the core architectural principles of relational modeling through graph attention and hybrid domain adaptation represent general design patterns that could potentially be adapted to other cross-domain learning problems. However, such extensions would require domain-specific graph construction strategies, appropriate connectivity patterns, and thorough empirical validation in each target domain. Future work will systematically explore this transferability by conducting preliminary experiments in medical imaging and action recognition, improving computational efficiency through graph sparsification, enhancing robustness on small or noisy datasets, and extending to multimodal emotion recognition. These investigations will help determine whether GAT-ADA’s design principles can serve as a general paradigm for domain-adaptive learning across diverse applications, or whether the observed success is primarily specific to the structured nature of facial expression data.

\newpage

\bibliography{acml25}

@article{DBLP:journals/cviu/GuoZ19,
  author       = {Guodong Guo and
                  Na Zhang},
  title        = {A survey on deep learning based face recognition},
  journal      = {Comput. Vis. Image Underst.},
  volume       = {189},
  year         = {2019},
  doi          = {10.1016/J.CVIU.2019.102805},
  bibsource    = {dblp computer science bibliography, https://dblp.org}
}

@article{DBLP:journals/cviu/JaimesS07,
  author       = {Alejandro Jaimes and
                  Nicu Sebe},
  title        = {Multimodal human-computer interaction: {A} survey},
  journal      = {Comput. Vis. Image Underst.},
  volume       = {108},
  number       = {1-2},
  pages        = {116--134},
  year         = {2007},
  doi          = {10.1016/J.CVIU.2006.10.019},
  bibsource    = {dblp computer science bibliography, https://dblp.org}
}

@article{DBLP:journals/tii/BisogniCHNU22,
  author       = {Carmen Bisogni and
                  Aniello Castiglione and
                  Sanoar Hossain and
                  Fabio Narducci and
                  Saiyed Umer},
  title        = {Impact of Deep Learning Approaches on Facial Expression Recognition
                  in Healthcare Industries},
  journal      = {{IEEE} Trans. Ind. Informatics},
  volume       = {18},
  number       = {8},
  pages        = {5619--5627},
  year         = {2022},
  doi          = {10.1109/TII.2022.3141400},
  bibsource    = {dblp computer science bibliography, https://dblp.org}
}

@article{DBLP:journals/inffus/PoriaCBH17,
  author       = {Soujanya Poria and
                  Erik Cambria and
                  Rajiv Bajpai and
                  Amir Hussain},
  title        = {A review of affective computing: From unimodal analysis to multimodal
                  fusion},
  journal      = {Inf. Fusion},
  volume       = {37},
  pages        = {98--125},
  year         = {2017},
  doi          = {10.1016/J.INFFUS.2017.02.003},
  bibsource    = {dblp computer science bibliography, https://dblp.org}
}

@article{DBLP:journals/corr/abs-2408-15777,
  author       = {Yan Wang and
                  Shaoqi Yan and
                  Yang Liu and
                  Wei Song and
                  Jing Liu and
                  Yang Chang and
                  Xinji Mai and
                  Xiping Hu and
                  Wenqiang Zhang and
                  Zhongxue Gan},
  title        = {A Survey on Facial Expression Recognition of Static and Dynamic Emotions},
  journal      = {CoRR},
  volume       = {abs/2408.15777},
  year         = {2024},
  doi          = {10.48550/ARXIV.2408.15777},
  eprinttype    = {arXiv},
  eprint       = {2408.15777},
  bibsource    = {dblp computer science bibliography, https://dblp.org}
}

@article{DBLP:journals/pami/ChenPWXLL22,
  author       = {Tianshui Chen and
                  Tao Pu and
                  Hefeng Wu and
                  Yuan Xie and
                  Lingbo Liu and
                  Liang Lin},
  title        = {Cross-Domain Facial Expression Recognition: {A} Unified Evaluation
                  Benchmark and Adversarial Graph Learning},
  journal      = {{IEEE} Trans. Pattern Anal. Mach. Intell.},
  volume       = {44},
  number       = {12},
  pages        = {9887--9903},
  year         = {2022},
  doi          = {10.1109/TPAMI.2021.3131222},
  bibsource    = {dblp computer science bibliography, https://dblp.org}
}

@article{DBLP:journals/pami/GopalanLC14,
  author       = {Raghuraman Gopalan and
                  Ruonan Li and
                  Rama Chellappa},
  title        = {Unsupervised Adaptation Across Domain Shifts by Generating Intermediate
                  Data Representations},
  journal      = {{IEEE} Trans. Pattern Anal. Mach. Intell.},
  volume       = {36},
  number       = {11},
  pages        = {2288--2302},
  year         = {2014},
  doi          = {10.1109/TPAMI.2013.249},
  bibsource    = {dblp computer science bibliography, https://dblp.org}
}

@article{DBLP:journals/taffco/LiD22,
  author       = {Shan Li and
                  Weihong Deng},
  title        = {A Deeper Look at Facial Expression Dataset Bias},
  journal      = {{IEEE} Trans. Affect. Comput.},
  volume       = {13},
  number       = {2},
  pages        = {881--893},
  year         = {2022},
  doi          = {10.1109/TAFFC.2020.2973158},
  bibsource    = {dblp computer science bibliography, https://dblp.org}
}

@article{DBLP:journals/tmm/GaoXHCL24,
  author       = {Yuefang Gao and
                  Yuhao Xie and
                  Zeke Zexi Hu and
                  Tianshui Chen and
                  Liang Lin},
  title        = {Adaptive Global-Local Representation Learning and Selection for Cross-Domain
                  Facial Expression Recognition},
  journal      = {{IEEE} Trans. Multim.},
  volume       = {26},
  pages        = {6676--6688},
  year         = {2024},
  doi          = {10.1109/TMM.2024.3355637},
  bibsource    = {dblp computer science bibliography, https://dblp.org}
}

@article{DBLP:journals/information/KopalidisSVD24,
  author       = {Thomas Kopalidis and
                  Vassilios Solachidis and
                  Nicholas Vretos and
                  Petros Daras},
  title        = {Advances in Facial Expression Recognition: {A} Survey of Methods,
                  Benchmarks, Models, and Datasets},
  journal      = {Inf.},
  volume       = {15},
  number       = {3},
  pages        = {135},
  year         = {2024},
  doi          = {10.3390/INFO15030135},
  bibsource    = {dblp computer science bibliography, https://dblp.org}
}

@inproceedings{DBLP:conf/iclr/VelickovicCCRLB18,
  author       = {Petar Veli{\v{c}}kovi{\'c} and
                  Guillem Cucurull and
                  Arantxa Casanova and
                  Adriana Romero and
                  Pietro Li{\`o} and
                  Yoshua Bengio},
  title        = {Graph Attention Networks},
  booktitle    = {6th International Conference on Learning Representations, {ICLR} 2018,
                  Vancouver, BC, Canada, April 30 – May 3, 2018, Conference Track Proceedings},
  year         = {2018},
  bibsource    = {dblp computer science bibliography, https://dblp.org}
}

@inproceedings{DBLP:conf/eccv/SunS16,
  author       = {Baochen Sun and
                  Kate Saenko},
  editor       = {Gang Hua and
                  Herv{\'{e}} J{\'{e}}gou},
  title        = {Deep {CORAL:} Correlation Alignment for Deep Domain Adaptation},
  booktitle    = {Computer Vision - {ECCV} 2016 Workshops - Amsterdam, The Netherlands,
                  October 8-10 and 15-16, 2016, Proceedings, Part {III}},
  series       = {Lecture Notes in Computer Science},
  volume       = {9915},
  pages        = {443--450},
  year         = {2016},
  doi          = {10.1007/978-3-319-49409-8\_35},
  bibsource    = {dblp computer science bibliography, https://dblp.org}
}

@article{DBLP:journals/jmlr/GrettonBRSS12,
  author       = {Arthur Gretton and
                  Karsten M. Borgwardt and
                  Malte J. Rasch and
                  Bernhard Sch{\"o}lkopf and
                  Alexander J. Smola},
  title        = {A Kernel Two-Sample Test},
  journal      = {J. Mach. Learn. Res.},
  volume       = {13},
  pages        = {723--773},
  year         = {2012},
  ee           = {https://dl.acm.org/doi/10.5555/2188385.2188410},
  bibsource    = {dblp computer science bibliography, https://dblp.org}
}

@article{DBLP:journals/jmlr/GaninUAGLLML16,
  author       = {Yaroslav Ganin and
                  Evgeniya Ustinova and
                  Hana Ajakan and
                  Pascal Germain and
                  Hugo Larochelle and
                  Fran{\c{c}}ois Laviolette and
                  Mario Marchand and
                  Victor S. Lempitsky},
  title        = {Domain-Adversarial Training of Neural Networks},
  journal      = {J. Mach. Learn. Res.},
  volume       = {17},
  number       = {59},
  pages        = {1--35},
  year         = {2016},
  bibsource    = {dblp computer science bibliography, https://dblp.org}
}

@inproceedings{DBLP:conf/cvpr/LuceyCKSAM10,
  author       = {Patrick Lucey and
                  Jeffrey F. Cohn and
                  Takeo Kanade and
                  Jason M. Saragih and
                  Zara Ambadar and
                  Iain A. Matthews},
  title        = {The Extended Cohn‑Kanade Dataset (CK+): A complete dataset for action unit and emotion‑specified expression},
  booktitle    = {2010 IEEE Computer Society Conference on Computer Vision and Pattern Recognition Workshops (CVPRW)},
  pages        = {94--101},
  year         = {2010},
  doi          = {10.1109/CVPRW.2010.5543262},
  bibsource    = {dblp computer science bibliography}
}

@inproceedings{DBLP:conf/cvpr/LiDD17,
  author       = {Shan Li and
                  Weihong Deng and
                  JunPing Du},
  title        = {Reliable Crowdsourcing and Deep Locality‑Preserving Learning for Expression Recognition in the Wild},
  booktitle    = {Proceedings of the IEEE Conference on Computer Vision and Pattern Recognition (CVPR)},
  pages        = {2852--2861},
  month        = {July},
  year         = {2017},
  bibsource    = {dblp computer science bibliography}
}

@misc{DBLP:dataset/jaffe/LyonsAKG98,
  author       = {Michael J. Lyons and
                  Shigeru Akamatsu and
                  Miyuki G. Kamachi and
                  Jiro Gyoba},
  title        = {The Japanese Female Facial Expression (JAFFE) Database},
  howpublished = {Website, originally presented alongside the paper “Coding Facial Expressions with Gabor Wavelets” at the Third IEEE International Conference on Automatic Face and Gesture Recognition, Nara, Japan, April 14–16,1998},
  year         = {1998},
}

@misc{DBLP:dataset/rafdb/LiDD17,
  author       = {Shan Li and
                  Weihong Deng and
                  JunPing Du},
  title        = {The Real-world Affective Faces Database (RAF‑DB)},
  howpublished = {Crowdsourced dataset from “Reliable Crowdsourcing and Deep Locality‑Preserving Learning for Expression Recognition in the Wild,” presented at CVPR2017},
  year         = {2017},
}

@misc{DBLP:dataset/fer2013/GoodfellowCSC13,
  author       = {Ian J. Goodfellow and
                  David Erhan and
                  Pierre-Luc Carrier and
                  Aaron Courville and
                  Yoshua Bengio},
  title        = {The Facial Expression Recognition 2013 (FER2013) Dataset},
  howpublished = {Introduced as part of "Challenges in Representation Learning: A report on three machine learning contests" at ICML2013},
  year         = {2013},
}

@misc{DBLP:dataset/sfew2/SFEW2.0,
  author       = {Abhinav Dhall and
                  Roland Goecke and
                  Simon Lucey and
                  Tom Gedeon},
  title        = {Static Facial Expressions in the Wild 2.0 (SFEW2.0) Dataset},
  howpublished = {Selected static frames from the AFEW video dataset as part of the EmotiW2015 challenge},
  year         = {2015},
}

@misc{DBLP:dataset/expw/ZhangLoy17,
  author       = {Zhanpeng Zhang and
                  Ping Luo and
                  Chen Change Loy and
                  Xiaoou Tang},
  title        = {Expression in-the-Wild (ExpW) Dataset},
  howpublished = {Introduced in “From Facial Expression Recognition to Interpersonal Relation Prediction,” Int. J. Comput. Vis. (2018)},
  year         = {2017},
}

@inproceedings{DBLP:conf/icb/ZhuSZ16,
  author    = {Ronghang Zhu and Gaoli Sang and Qijun Zhao},
  title     = {Discriminative Feature Adaptation for Cross-Domain Facial Expression Recognition},
  booktitle = {2016 International Conference on Biometrics (ICB)},
  pages     = {1--7},
  year      = {2016},
  bibsource = {dblp computer science bibliography}
}

@inproceedings{DBLP:conf/sibgrapi/ZavarezBO17,
  author       = {Marcus Vinicius Zavarez and
                  Rodrigo Ferreira Berriel and
                  Thiago Oliveira{-}Santos},
  title        = {Cross-Database Facial Expression Recognition Based on Fine-Tuned Deep Convolutional Network},
  booktitle    = {2017 30th SIBGRAPI Conference on Graphics, Patterns and Images (SIBGRAPI)},
  pages        = {405--412},
  year         = {2017},
  organization = {{IEEE}},
  bibsource    = {dblp computer science bibliography, https://dblp.org}
}

@inproceedings{DBLP:conf/icpr/0001D18,
  author       = {Shan Li and
                  Weihong Deng},
  title        = {Deep Emotion Transfer Network for Cross-database Facial Expression Recognition},
  booktitle    = {2018 24th International Conference on Pattern Recognition (ICPR)},
  pages        = {3092--3099},
  year         = {2018},
  doi          = {10.1109/ICPR.2018.8545284},
  bibsource    = {dblp computer science bibliography, https://dblp.org}
}

@inproceedings{DBLP:conf/nips/LongCWCJ18,
  author       = {Mingsheng Long and
                  Zhangjie Cao and
                  Jianmin Wang and
                  Michael I. Jordan},
  title        = {Conditional Adversarial Domain Adaptation},
  booktitle    = {Advances in Neural Information Processing Systems},
  volume       = {31},
  year         = {2018},
  pages        = {1647--1657},
  doi          = {10.5555/3326943.3327094},
  bibsource    = {dblp computer science bibliography, https://dblp.org}
}

@article{DBLP:journals/neucom/JiHYSS19,
  author       = {Yanli Ji and
                  Yuhan Hu and
                  Yang Yang and
                  Fumin Shen and
                  Heng Tao Shen},
  title        = {Cross-Domain Facial Expression Recognition via an Intra-Category Common Feature and Inter-Category Distinction Feature Fusion Network},
  journal      = {Neurocomputing},
  volume       = {333},
  pages        = {231--239},
  year         = {2019},
  doi          = {10.1016/j.neucom.2018.12.037},
  bibsource    = {dblp computer science bibliography, https://dblp.org}
}

@inproceedings{DBLP:conf/iccv/XuLYL19,
  author       = {Ruijia Xu and
                  Guanbin Li and
                  Jihan Yang and
                  Liang Lin},
  title        = {Larger Norm More Transferable: An Adaptive Feature Norm Approach for Unsupervised Domain Adaptation},
  booktitle    = {2019 {IEEE/CVF} International Conference on Computer Vision (ICCV)},
  pages        = {1426--1435},
  year         = {2019},
  doi          = {10.1109/ICCV.2019.00151},
  bibsource    = {dblp computer science bibliography, https://dblp.org}
}

@inproceedings{DBLP:conf/cvpr/LiZLGR20,
  author       = {Mengxue Li and
                  Yi‑Ming Zhai and
                  You‑Wei Luo and
                  Peng‑Fei Ge and
                  Chuan‑Xian Ren},
  title        = {Enhanced Transport Distance for Unsupervised Domain Adaptation},
  booktitle    = {2020 IEEE/CVF Conference on Computer Vision and Pattern Recognition (CVPR)},
  pages        = {13936--13944},
  year         = {2020},
  doi          = {10.1109/CVPR42600.2020.01395},
  bibsource    = {dblp computer science bibliography, https://dblp.org}
}

@article{DBLP:journals/taffc/LiD22,
  author       = {Shan Li and
                  Weihong Deng},
  title        = {A Deeper Look at Facial Expression Dataset Bias},
  journal      = {{IEEE} Transactions on Affective Computing},
  volume       = {13},
  number       = {2},
  pages        = {881--893},
  year         = {2022},
  doi          = {10.1109/TAFFC.2020.2973158},
  bibsource    = {dblp computer science bibliography, https://dblp.org}
}

@inproceedings{DBLP:conf/icml/FatrasSFC21,
  author       = {Kilian Fatras and
                  Thibault Séjourné and
                  Rémi Flamary and
                  Nicolas Courty},
  title        = {Unbalanced Minibatch Optimal Transport: Applications to Domain Adaptation},
  booktitle    = {Proceedings of the 38th International Conference on Machine Learning (ICML)},
  pages        = {3186--3197},
  year         = {2021},
  bibsource    = {dblp computer science bibliography, https://dblp.org}
}

@article{DBLP:journals/tmm/LiZCLZ22,
  author       = {Yingjian Li and
                  Zheng Zhang and
                  Bingzhi Chen and
                  Guangming Lu and
                  David Zhang},
  title        = {Deep Margin‑Sensitive Representation Learning for Cross‑Domain Facial Expression Recognition},
  journal      = {{IEEE} Transactions on Multimedia},
  volume       = {25},
  pages        = {1359--1373},
  year         = {2022},
  doi          = {10.1109/TMM.2022.3141604},
  bibsource    = {dblp computer science bibliography, https://dblp.org}
}

@article{DBLP:journals/peerjcs/AlzahraniAASIASA25,
  author       = {Ahmed Omar Alzahrani and
                  Ahmed Mohammed Alghamdi and
                  M. Usman Ashraf and
                  Iqra Ilyas and
                  Nadeem Sarwar and
                  Abdulrahman Alzahrani and
                  Alaa Abdul Salam Alarood},
  title        = {A Novel Facial Expression Recognition Framework Using Deep Learning Based Dynamic Cross-Domain Dual Attention Network},
  journal      = {PeerJ Comput. Sci.},
  volume       = {11},
  pages        = {e2866},
  year         = {2025},
  publisher    = {PeerJ Inc.},
  url          = {},
  doi          = {10.7717/peerj-cs.2866},
  bibsource    = {dblp computer science bibliography, https://dblp.org}
}

@article{DBLP:journals/corr/abs-1609-02907,
  author       = {Thomas N. Kipf and
                  Max Welling},
  title        = {Semi-Supervised Classification with Graph Convolutional Networks},
  journal      = {CoRR},
  volume       = {abs/1609.02907},
  year         = {2016},
  url          = {},
  eprinttype   = {arXiv},
  eprint       = {1609.02907},
  bibsource    = {dblp computer science bibliography, https://dblp.org}
}

@article{DBLP:journals/mta/GuoWLLZW24,
  author       = {Zhe Guo and
                  Bingxin Wei and
                  Jiayi Liu and
                  Xuewen Liu and
                  Zhibo Zhang and
                  Yi Wang},
  title        = {{USTST:} unsupervised self-training similarity transfer for cross-domain
                  facial expression recognition},
  journal      = {Multim. Tools Appl.},
  volume       = {83},
  number       = {14},
  pages        = {41703--41723},
  year         = {2024},
  url          = {},
  doi          = {10.1007/s11042-023-17362-9},
  bibsource    = {dblp computer science bibliography, https://dblp.org}
}

@article{DBLP:journals/eaai/ZhuAXWYJH25,
  author       = {Yanan Zhu and
                  Jiaqiu Ai and
                  Weibao Xue and
                  Mingyang Wu and
                  Sen Yang and
                  Wei Jia and
                  Min Hu},
  title        = {Cross-domain facial expression recognition: {Bi-Directional} Fusion
                  of Active and Stable Information},
  journal      = {Eng. Appl. Artif. Intell.},
  volume       = {149},
  pages        = {110357},
  year         = {2025},
  url          = {},
  doi          = {10.1016/j.engappai.2024.110357},
  bibsource    = {dblp computer science bibliography, https://dblp.org}
}

@inproceedings{DBLP:conf/bmvc/Prados-TorreblancaBB22,
  author    = {Andr{\'e}s Prados{-}Torreblanca and
               Jos{\'e} M. Buenaposada and
               Luis Baumela},
  title     = {Shape Preserving Facial Landmarks with Graph Attention Networks},
  booktitle = {British Machine Vision Conference (BMVC)},
  year      = {2022}
}

@inproceedings{DBLP:conf/fgr/MaM23,
  author    = {Guanming Ma and Huazhu Ma},
  title     = {Relation-aware Network for Facial Expression Recognition},
  booktitle = {17th IEEE International Conference on Automatic Face and Gesture Recognition (FG)},
  year      = {2023}
}
\end{document}